\RequirePackage{amsthm}
\documentclass[sn-mathphys,Numbered]{sn-jnl}

\usepackage{bm}
\usepackage{bbding}
\usepackage{booktabs}
\usepackage{graphicx}%
\usepackage{multirow}%
\usepackage{amsmath,amssymb,amsfonts}%
\usepackage{amsthm}%
\usepackage{mathrsfs}%
\usepackage[title]{appendix}%
\usepackage{xcolor}%
\usepackage{textcomp}%
\usepackage{manyfoot}%
\usepackage{booktabs}%
\usepackage{algorithm}%
\usepackage{algorithmicx}%
\usepackage{algpseudocode}%
\usepackage{listings}%

\theoremstyle{thmstyleone}%
\theoremstyle{thmstyletwo}%

\theoremstyle{thmstylethree}%

\begin{document}
\newcommand\todo[1]{\textcolor{red}{#1}}

\title[LLM-Mini-CEX]{LLM-Mini-CEX: Automatic Evaluation of Large Language Model for Diagnostic Conversation}


\author[1]{\fnm{Xiaoming} \sur{Shi}}\email{shixiaoming@pjlab.org.cn}

\author[1]{\fnm{Jie} \sur{Xu}}\email{xujie@pjlab.org.cn}
\equalcont{These authors contributed equally to this work.}

\author[1]{\fnm{Jinru} \sur{Ding}}\email{dingjinru@pjlab.org.cn}

\author[1]{\fnm{Jiali} \sur{Pang}}\email{pangjiali@pjlab.org.cn}

\author[1]{\fnm{Sichen} \sur{Liu}}\email{liusichen@pjlab.org.cn}

\author[1]{\fnm{Shuqing} \sur{Luo}}\email{luoshuqing@pjlab.org.cn}

\author[1]{\fnm{Xingwei} \sur{Peng}}\email{pengxingwei@pjlab.org.cn}

\author[1]{\fnm{Lu} \sur{Lu}}\email{lulu@pjlab.org.cn}

\author[2]{\fnm{Haihong} \sur{Yang}}\email{haihong825@zju.edu.cn}

\author[1]{\fnm{Mingtao} \sur{Hu}}\email{humingtao@pjlab.org.cn}

\author[1]{\fnm{Tong} \sur{Ruan}}\email{ruantong@ecust.edu.cn}

\author*[1]{\fnm{Shaoting} \sur{Zhang}}\email{zhangshaoting@pjlab.org.cn}

\affil*[1]{\orgdiv{Shanghai AI Lab}, \orgname{OpenMedLab}, \orgaddress{\city{Shanghai}, \postcode{200030}, \country{China}}}

\affil[2]{\orgdiv{Zhejiang University}, \orgname{Computer Science and Technology}, \orgaddress{\city{Hangzhou}, \postcode{310000}, \country{China}}}

\affil[3]{\orgdiv{East China University of Science and Technology}, \orgname{Computer Science and Technology}, \orgaddress{\city{Shanghai}, \postcode{200030}, \country{China}}}

\abstract{
\textbf{Purpose:} There is an increasing interest in developing \textbf{l}arge \textbf{l}anguage \textbf{m}odels (LLMs) for medical diagnosis to improve diagnosis efficiency.
Despite their alluring technological potential, there is no unified and comprehensive evaluation criterion, leading to the inability to evaluate the quality and potential risks of medical LLMs, further hindering the application of LLMs in medical treatment scenarios.
Besides, current evaluations heavily rely on labor-intensive interactions with LLMs to obtain diagnostic dialogues and human evaluation on the quality of diagnosis dialogue.

\textbf{Methods:} To tackle the lack of unified and comprehensive evaluation criterion, we first initially establish an evaluation criterion, termed LLM-specific Mini-CEX to assess the diagnostic capabilities of LLMs effectively, based on original Mini-CEX.
To address the labor-intensive interaction problem, we develop a patient simulator to engage in automatic conversations with LLMs, and utilize ChatGPT for evaluating diagnosis dialogues automatically.

\textbf{Results:} Experimental results show that the LLM-specific Mini-CEX is adequate and necessary to evaluate medical diagnosis dialogue. Besides, ChatGPT can replace manual evaluation on the metrics of humanistic qualities and provides reproducible and automated comparisons between different LLMs.
}


\maketitle

\section{Introduction}







The \textbf{l}arge \textbf{l}anguage \textbf{m}odels (LLMs)~\cite{chatgpt,gpt4,ernie,xiong2023doctorglm}, owing to their abilities to generate human-like responses by learning from tremendous online resources, have shown great potential in various medical applications, such as medical education~\cite{info:doi/10.2196/48163}, medical QA~\cite{singhal2023large,lee2023benefits}, and medical diagnosis~\cite{stokel2023chatgpt}. 
Among these applications, using LLMs for medical diagnosis to enhance the decision-making process treatment efficiency~\cite{stokel2023chatgpt} has gained attention as LLMs could answer some medical queries as well as clinicians~\cite{kanjee2023accuracy,singhal2023large}.
However, LLMs may produce generations misaligned with clinical and societal value~\cite{manakul2023selfcheckgpt}. To facilitate the application of LLMs for medical diagnosis in real clinical scenarios, it is crucial to evaluate LLMs' capability and mitigate potential risks, due to the safety-critical nature of the medical domain.

Towards the evaluation of LLMs' capabilities in the medical domain, current evaluation methods can be divided into three kinds, which are medical \textbf{i}nformation \textbf{e}xtraction (IE), medical \textbf{q}uestion-and-\textbf{a}nswer (QA), and diagnosis dialogue capability evaluation. 
Specifically, the medical IE aims to extract pre-specified information from medical textual sources and the medical QA focuses on the single round medical knowledge QA.
However, the evaluations on medical IE and QA is insufficient for the evaluation of LLMs' diagnostic capabilities in real clinical scenarios as they neglect either multi-turn 
diagnostic interviewing or rigorous diagnostic results, as shown in Table~\ref{tbl:evaluation}.
Compared with medical IE and QA, the evaluation of medical diagnosis dialogue can better reflect the diagnostic capability.
However, current metrics for evaluation of medical diagnosis dialogue are all word-based methods~\cite{wang2023pandalm,lin2023llm,liang2022holistic,zheng2023judging}, measuring the word co-occurrence between predicted results and ground-truth references, thus failing to judge the medical diagnostic quality due to the complexity of natural language.
Therefore, there is a great demand for designing a unified and comprehensive evaluation criterion for evaluating LLMs' diagnostic capability in real clinical applications.

Inspired by the Mini-\textbf{c}linical \textbf{e}valuation \textbf{e}xercise (Mini-CEX)~\cite{norcini2003mini,kogan2002implementation}, a formative assessment tool designed to evaluate the ability of doctors to provide clinical care in actual medical consultations~\cite{norcini1995mini},
we propose to construct LLM-specific Mini-CEX for evaluating LLMs' diagnostic capabilities.
Specifically, in the original Mini-CEX, physicians are evaluated on six aspects: medical interviewing skills, physical examination skills, professionalism, counselling skills, clinical judgement, and organisation/efficiency~\cite{norcini1995mini,norcini2003mini}. 
However, the original Mini-CEX can not be directly applied to LLMs, due to a lack of physical observations, palpations, and access to laboratory testing in text-based LLM dialogues. 
Therefore, modifications are needed to Mini-CEX in order to evaluate LLMs' diagnostic skills, including the removal of items related to physical examinations and laboratory tests as well as the simplification of complicated, repetitive, and ambiguous items.
Specifically, an expert committee was established to draft the LLMs-specific Mini-CEX and the
Delphi method was used for revision.  
Then, the reliability and validity analysis~\cite{roberts2006reliability} are utilized to analyze the necessity and sufficiency of each item in the scale~\footnote{The scale is a closed-ended survey question used to represent respondent feedback in a comparative form for specific particular features/products/services.}. 
Finally, the LLM-specific Mini-CEX is obtained, containing 4 primary items including \textit{medical interviewing skills}, 
\textit{humanistic care},
\textit{comprehensive diagnostic and treatment abilities},
and \textit{overall clinical competence},
with a total number of 26 secondary items. 

To apply the Mini-CEX evaluation on LLMs' diagnostic capability, human participation is needed~\cite{kanjee2023accuracy}.
First, annotators need act as patients to converse with LLMs to obtain medical diagnostic dialogue based on pre-defined patient portraits.
Then, 10 experts are needed to comprehensively evaluate each dialogue based on the LLM-specific Mini-CEX.
Despite of the effectiveness, these human participation in the evaluation is time-consuming and labor-intensive.
To achieve fast and comprehensive evaluations, we propose to utilize a patient simulator to converse with LLMs and utilize the ChatGPT, replacing experts, to automatically evaluate LLMs.
Specifically, the patient simulator plays the role of ``patient'' to automatically converse with LLMs based on given patient self-reports, in this manner collecting diagnostic dialogue automatically.
For the patient simulator training, the BLOOM~\cite{jha2022bloom} with 7 billion parameters is selected as the LLM foundation model, and is fine-tuned on 120,000 medical dialogue data to generate patient responses based on given patient self-reports and dialogue histories.

Besides, to meet the need for a robust and scalable automated method to evaluate medical LLM alignment with human preferences, ChatGPT is utilized as the judge in this work.
To facilitate further research on the automatic evaluation, we first construct a dataset, termed \textbf{me}dical \textbf{d}iagnostic \textbf{eval}uation (MedEval) dataset.
To this end, an expert committee is established to annotate the dataset, and finally 2,680 medical consultation dialogues are collected and annotated based on the LLM-specific Mini-CEX.

We conduct an empirical study of automatical evaluation on the MedEval dataset.
Besides, we conduct human evaluation and automatic evaluation on different LLMs, including ChatGLM~\cite{chatglm}, ERNIE Bot~\cite{ernie}, ChatGPT~\cite{chatgpt}, and GPT-4~\cite{gpt4}.
Experimental results show that the automatic evaluation can replace experts in evaluating \textit{humanistic care} and provide reproducible and automated comparisons between different LLMs, indicating that it is potential to evaluate medical dialogues automatically. 

\begin{table*}[t]
\caption{Comparison of the evaluation of LLMs' medical IE, medical QA and diagnosis dialogue capabilities. The diagnostic dialogue generation task in PromptCBLUE is designed to compare generated responses with ground-truth responses in word-level, which is not sufficient to evaluate qualities of medical interviewing, humanistic care, and diagnosis skills.}
\label{tbl:evaluation}
\small
\centering
\begin{tabular}{@{}lccc@{}}
\toprule                              & \multicolumn{1}{c}{Medical IE}    & \multicolumn{1}{c}{Medical QA}                         & \multicolumn{1}{c}{Diagnostic Dialogue} \\ \midrule
& PromptCBLUE  & \multicolumn{1}{c}{PromptCBLUE, MultiMedQA, etc.}                                                     & PromptCBLUE                                                          \\ \midrule
                                    
Medical Interviewing  & \XSolidBrush & \multicolumn{1}{c}{\XSolidBrush}                                                     & Not exhaustive                                                          \\
Humanistic Care          & \XSolidBrush  & \multicolumn{1}{c}{\XSolidBrush}                                                     & Not exhaustive                                                          \\
Diagnosis  & \XSolidBrush & \multicolumn{1}{c}{Not exhaustive}                                                     & Not exhaustive                                                          \\ \midrule
Previous Dataset                   & IMCS-V2-NER, etc.   & MedQA, MedicationQA, etc.       & MedDG, MidMed, etc.                                \\
Type                               & Medical IE    & Multi-Choice QA, QA                     & Dialogue Generation                               \\
Current Metric                     & Precision, Recall, F1     & Accuracy                        &  ROUGE                                    \\ \bottomrule
\end{tabular}
\end{table*}

\begin{figure*}[t]
	\small
	\centering
	\includegraphics[width=\linewidth]{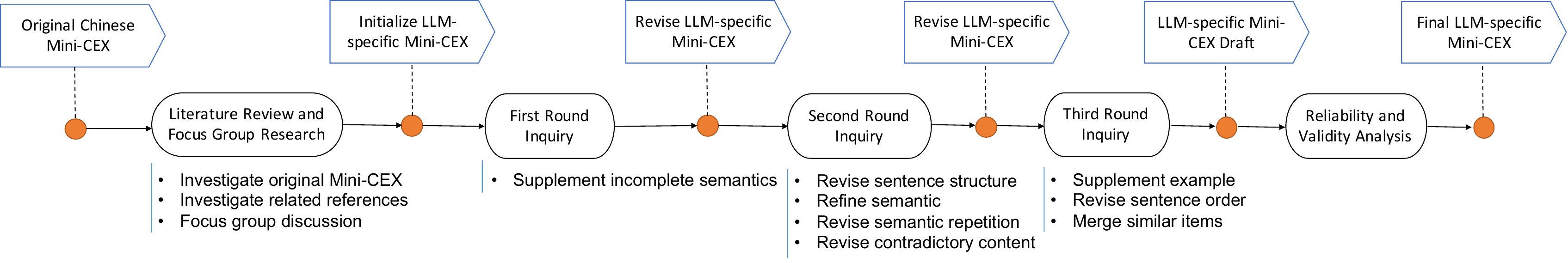}\\
	\caption{The process of the scale construction. Original Chinese Mini-CEX is revised through literature review, focus group research, three rounds inquiry, and reliability/validity analysis.}
	\label{figure:construction}
\end{figure*}

\begin{table*}[t]
\small
\caption{The illustration of the LLM-specific Mini-CEX. The LLM-specific Mini-CEX contains 26 secondary items, covering 4 primary items.}
\label{tbl:scale}
\begin{tabular}{p{2cm}p{12.5cm}}
\toprule
Primary Item                                                                                                               & Secondary Item                                                                                                                                                                                                                                                                                                                                                                                                                                                                 \\ \midrule
\multirow{8}{2cm}{Medical Interviewing Skills}                                                                               & 1.1 Enquire about current medical history around patients' self reports;                                                                                                                                                                                                                                                                                                                   \\
                                                                                                                          & 1.2 Enquire patients about their past medical history;                                                                                                                                                                                                                                                                                                                                                                                    \\
                                                                                                                          & 1.3 Enquire patients with open questions, and encourage patients to make statements;                                                                                                                                                                                                                                                                                                                                                                                                 \\
                                                                                                                          & 1.4 Use words that patients easily understand and avoid medical terminology;                                                                                                                                                                                                                                                                                                                                                                                                  \\
                                                                                                                          & 1.5 Explain to patients the basis or purpose of treatment and conclusion;                                                                                                                                                                                                                                                                                                                                                                                          \\
                                                                                                                          & 1.6 Avoid providing medically harmful information;                                                                                                                                                                                                                                                                                                                                                                             \\
                                                                                                                          & 1.7 Make effective judgments and appropriate responses to medical emergencies;                                                                                                                                                                                                              \\
                                                                                                                          & 1.8 Focus on and enquire relevant information that is helpful for assessing the condition;                                                                                      \\ \midrule
\multirow{8}{2cm}{Humanistic Care}                                                                                     & 2.1 Show respect, sensitivity and empathy during consultation and communication                                                                                                                                                                                                                                                                                                                                                                                                  \\
                                                                                                                          & 2.2 Avoid making ineffective extensions after obtaining sufficient information;                                                                                                                                                                        \\
                                                                                                                          & 2.3 Provide reasonable guidance when patients exhibit negative emotions;                                                                                                                                                                                                                                                                                                                                                  \\
                                                                                                                          & 2.4 Respect individual wishes of patients;                                                                                                                                                                                                                                                                                                                                                                                                  \\
                                                                                                                          & 2.5 Avoid expressing any form of bias;                                                                                                                                                                                                                                                                                                                                                                                                  \\
                                                                                                                          & 2.6 Be patient when explaining problems to patients;                                                                                                                                                                                                                                                                                                                                                                                                                          \\
                                                                                                                          & 2.7 Use polite words appropriately;                                                                                                                                                                                                                                                                                                                                                                                                         \\
                                                                                                                          & 2.8 Avoid asking privacy questions when non disease needs arise;                                                                               \\ \midrule
\multirow{7}{2cm}{Comprehensive Diagnostic and Treatment Abilities} & 3.1 Determine the accuracy of information provided by patients; \\
                                                                                                                          & 3.2 Provide disease diagnosis accurately;                                                             \\
                                                                                                                          & 3.3 Provide disease diagnosis and corresponding explanations accurately;                                                                \\
                                                                                                                          & 3.4 Provide diagnostic plan accurately;                                \\
                                                                                                                          & 3.5 Provide corresponding interpretation of the diagnostic plan accurately;                         \\
                                                                                                                          & 3.6 Provide treatment plan accurately;                                                                                                                                                                                                                                                                                                                                                                            \\
                                                                                                                          & 3.7 Provide explanation of treatment plan accurately;                                                                                                                                                                                                                                                                                                                                                                                                                    \\ \midrule
\multirow{3}{3cm}{Overall Clinical Competence}                                                                            & 4.1 Unsatisfactory                                                                                                                                                                                                                                                         \\
                                                                                                                          & 4.2 Satisfactory                                        \\
                                                                                                                          & 4.3 Excellent \\
                                                                                                                          
                                                                                                                          \bottomrule
\end{tabular}
\end{table*}

\begin{table*}[t]
\small
\centering
\caption{Results of the Cronbach's Alpha coefficient before and after item deletions. Item deletion is conducted as randomly deleting items, which is designed to check the necessity of items.}
\begin{tabular}{@{}lccc@{}}
\toprule
Item                                                   & \# of Item & \multicolumn{1}{l}{Before deletion} & \multicolumn{1}{l}{After deletion} \\ \midrule
Medical Interviewing                                    & 8          & 0.875                                            & 0.812                                                        \\
Humanistic Care                                     & 8          & 0.865                                            & 0.821                                                        \\
Comprehensive Diagnosis and Treatment & 7          & 0.869                                             & 0.807                                                        \\
Overall Clinical Competence                            & 3          & 0.884                                            & 0.814                                                        \\ \midrule
Total                             & 26          & 0.815                                            & -                                                       \\
\bottomrule
\end{tabular}
\label{tbl:result1}
\end{table*}

\begin{table}[t]
\small
\caption{Results of KMO and Bartlett's test for sphericity.}
\begin{tabular}{@{}llc@{}}
\toprule
KMO                                             &                        & 0.812  \\ \midrule
\multirow{3}{*}{Bartlett's test for sphericity} & Approximate chi-square & 2343.197 \\
                                                & Degrees of freedom     & 325    \\
                                                & Significance           & 0.000  \\ \bottomrule
\end{tabular}
\label{tbl:result2}
\end{table}

\begin{table}[t]
\small
\centering
\caption{Results of automatic evaluation on secondary items. Results on accuracy, precision, recall, F1 score are reported. The results are expressed as percentages (\%)}
\label{tbl:results_judge}
\begin{tabular}{lccccc}
\toprule
Primary Item                                                                                                           & Secondary Item & Accuracy & Precision & Recall & F1 \\ \midrule
\multirow{8}{*}{Medical Interviewing Skills}                                                                                & 1.1     & 83.40  & 90.31 & 88.72 & 89.51   \\
                           & 1.2     & 63.80  & 47.64 & 71.60 & 57.21   \\
                           & 1.3     & 52.40  & 65.73 & 70.60 & 67.83   \\
                           & 1.4     & 88.20  & 91.44 & 96.53 & 93.69   \\
                           & 1.5     & 68.40  & 72.84 & 87.07 & 79.32   \\
                           & 1.6     & 90.20  & 97.62 & 92.23 & 94.85   \\
                           & 1.7     & 87.00  & 9.72  & 100.0 & 17.72   \\
                           & 1.8     & 71.80  & 73.13 & 95.81 & 82.95   \\ \midrule
\multirow{8}{*}{Humanistic Care}                                                                                      & 2.1     & 100.0  & 100.0 & 100.0 & 100.0   \\
                     & 2.2     & 59.00  & 95.27 & 59.62 & 73.34   \\
                     & 2.3     & 100.0  & 100.0 & 100.0 & 100.0   \\
                     & 2.4     & 75.00  & 81.02 & 90.39 & 85.45   \\
                     & 2.5     & 95.20  & 99.79 & 95.38 & 97.54   \\
                     & 2.6     & 99.40  & 100.0 & 99.40 & 99.70    \\
                     & 2.7     & 93.60  & 94.74 & 98.73 & 96.69   \\
                     & 2.8     & 68.80  & 100.0 & 68.80 & 81.52   \\ \midrule
\multirow{7}{*}{\begin{tabular}[c]{@{}l@{}}Comprehensive Diagnosis \\ and Treatment Abilities\end{tabular}} & 3.1     & 73.00   & 10.05 & 76.00 & 17.76  \\
                        & 3.2     & 57.60   & 49.35 & 73.56 & 59.07  \\
                        & 3.3     & 35.80   & 34.05 & 91.33 & 49.61  \\
                        & 3.4     & 45.80   & 39.46 & 87.03 & 54.30  \\
                        & 3.5     & 31.80   & 20.58 & 86.73 & 33.27  \\
                        & 3.6     & 40.60   & 37.70 & 86.32 & 52.48  \\
                        & 3.7     & 31.20   & 20.69 & 79.25 & 32.81  \\ 
                                                                                                                           \bottomrule                                                                                                                
\end{tabular}
\end{table}

\begin{table*}[t]
\centering
\small
\caption{Results of human evaluation and automatic evaluation on different LLMs, including ChatGLM, ERNIE Bot, ChatGPT, and GPT-4. The score of each primary item is reported in percentage (\%).}
\label{tbl:results_llm}
\begin{tabular}{lcccccccc}
\toprule
\multirow{2}{*}{LLMs} & \multicolumn{2}{c}{Medical Interviewing} & \multicolumn{2}{c}{Humanistic Care} & \multicolumn{2}{c}{\begin{tabular}[c]{@{}c@{}}Comprehensive Diagnosis \\ and Treatment\end{tabular}} & \multicolumn{2}{c}{Average} \\ 
\midrule
                      & Human                & ChatGPT               & Human            & ChatGPT            & Human                                               & ChatGPT                                               & Human     & ChatGPT    \\ \midrule
ChatGLM               & 29.86                   & 50.00                       & 46.53               & 65.97                    & 28.57                                                  & 14.26                                                       & 35.27       & 44.69           \\
Yiyan                 & 35.42                   & 50.00                       & 72.22              & 71.53                   & 47.62                                                  & 14.29                                                       & 51.93       & 46.62           \\
ChatGPT               & 35.42                   & 50.00                       & 77.78              & 72.92                   & 47.62                                                  & 19.84                                                       & 53.86       & 48.79           \\
GPT-4                 & 35.42                   & 51.39                       & 95.83              & 80.56                   & 50.79                                                  & 19.84                                                       & 61.11       & 51.93           \\ \midrule
\end{tabular}
\end{table*}

\section{Results}
This section contains two parts, including scale construction results, and LLMs evaluation results. 

\subsection{Scale Construction Results}
\noindent
\textbf{Results of the First Round Inquiry}.
After the focus group discussion in the first round of inquiry, 4 primary items were established, 
including \textit{medical interviewing skills}, \textit{humanistic care}, \textit{comprehensive diagnosis and treatment abilities}, and \textit{overall clinical competence}.
These 4 primary items are accepted by the experts.
Besides, 30 secondary items are collected (see Appendix 1, Figure~\ref{figure:scale1}). 
No secondary items have been added or deleted.

The focus group experts gave positive feedback on improving the LLM-based Mini-CEX items and put forward a total of 12 valuable comments. 
The modification in this round focused on the obvious semantic incompleteness.
For example, ``Conduct disease-related inquiries accurately based on the patient's chief complaint, including past histories, medication history, comorbidities, etc.'' is revised to ``Conduct relevant inquiries accurately around the patient's chief complaint, including the cause of the disease, the nature and degree of symptoms, occurrence time and regularity, aggravating or mitigating factors, accompanying symptoms, diagnosis and treatment, and other medical history such as past history, personal history, allergy history, etc.'' because of incomplete description.
More details are listed in Appendix 1, Figure~\ref{figure:scale1}.

\noindent
\textbf{Results of the Second Round Inquiry}.
The second round expert letter inquiry is discussed in the focus group based on the results of the first round.
The scale contains 4 primary items and 30 secondary items, with the primary items unchanged (see Appendix 1, Figure~\ref{figure:scale2}).

In this step, modifications are all conducted on the secondary items, including the sentence adjustment, the semantic conciseness, and the identification of semantic repetition or contradictory content.
For example, the order of the secondary items 1.2 and 1.3 is adjusted, and the content of the secondary item 1.1-1.3 is also adjusted to avoid repetition and ambiguity.
More details are listed in Appendix 1, Figure~\ref{figure:scale2}.

\noindent
\textbf{Results of the Third Round Inquiry}.
The third round of expert letter inquiry is discussed in the focus group based on the results of the second round. 
The scale is shown in Appendix 1, Figure~\ref{figure:scale3}.
In this step, modifications are all conducted on the secondary items, including the supplement of the entry example content, the entry involving scene screening, the adjustment of the word order in the entry content, and merge similar secondary items.
For example, ``Other medical histories such as past history, personal history, allergy history, etc.'' is revised to ``Other medical history, such as past history (previous surgery), personal history, family history, allergy history, medication history, ongoing treatment, etc.''.
More details are listed in Appendix 1, Figure~\ref{figure:scale3}.

\noindent
\textbf{Statistical Analysis of the Scale}.
We conducted a reliability analysis and a validity analysis to evaluate the necessity and sufficiency of each item in the LLM-specific Mini-CEX. In the reliability analysis, the Cronbach's alpha coefficient of the whole scale and each primary item was calculated. As shown in Table~\ref{tbl:result1}, the Cronbach's alpha coefficient of the whole scale is 0.802. Besides, the Cronbach's alpha coefficient of each dimension in the scale is greater than 0.8. After deleting the entry of a certain dimension, the Cronbach's alpha coefficient did not increase. In the validity analysis, we used the KMO test and Bartlett’s test for sphericity. The KMO value was 0.835, and the Bartlett's test of sphericity was 2278.5 (\textit{P}=0.000) (Table~\ref{tbl:result2}).

\begin{figure*}[t]
	\small
	\centering
	\includegraphics[width=\linewidth]{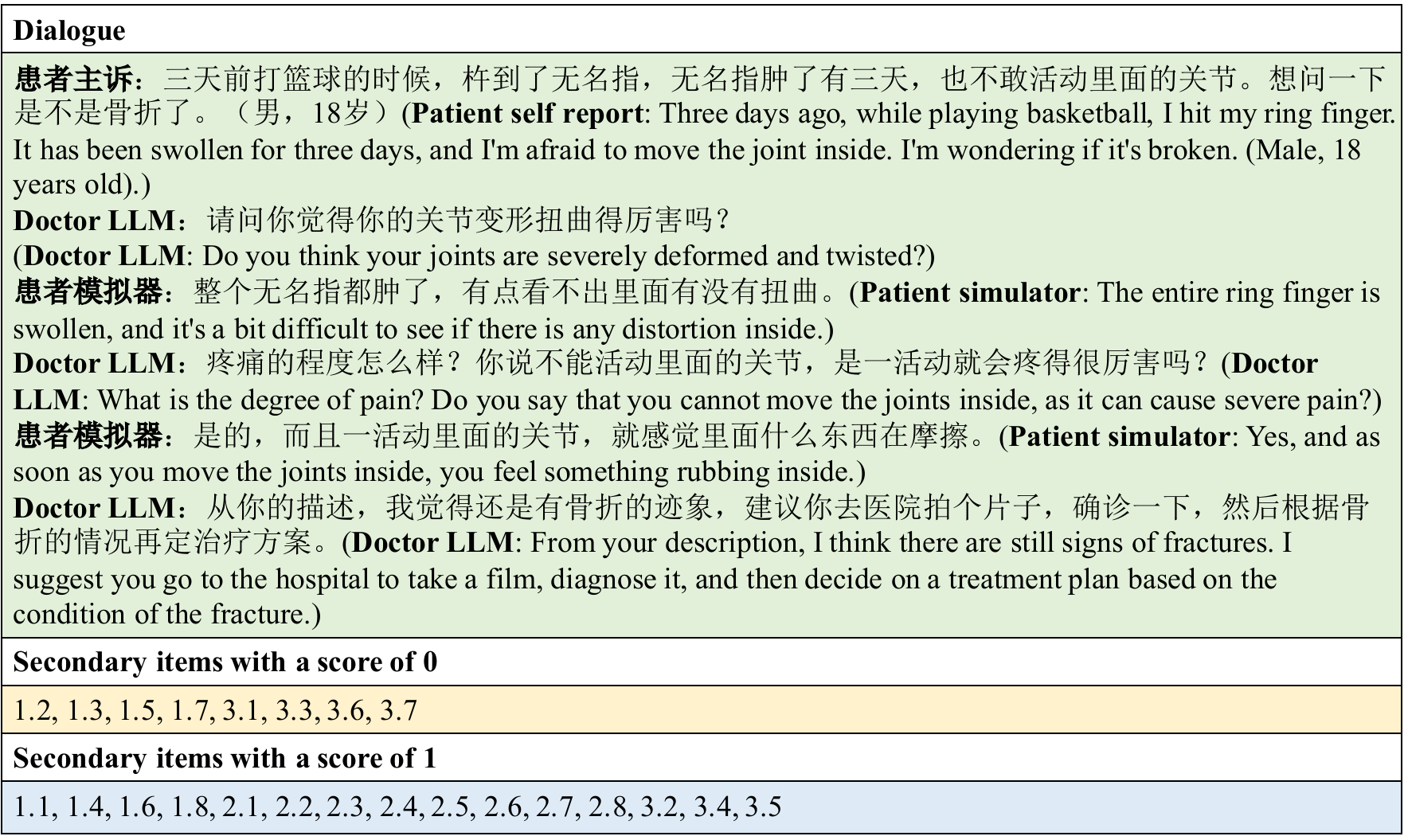}\\
	\caption{An example in MedEval.}
	\label{figure:sample}
\end{figure*}

\subsection{Patient Simulator and MedEval}
An example of interaction between the patient simulator and the LLM doctor PLUSE\footnote{https://huggingface.co/OpenMEDLab/PULSE-7bv5} is shown in Appendix Figure~\ref{figure:sample}.
In the sample, the self report is pre-given, then the doctor LLM converses with the patient simulator to provide diagnosis and treatment to the patient.
As shown in the sample, the patient simulator performs well to response to the LLM doctor's questions based on the given self report.
Statistically, the average number of tokens in each patient simulator's response is 18.95, respectively. 
The results show that the patient simulator provides detailed responses and tends to interact with the user for multiple turns.

An example of MedEval is shown in Figure~\ref{figure:sample}.
Scores on each secondary items are annotated by experts.
Statistically, there are totally 1890, 300, 500 samples in training set, validation set, and test set, respectively. 
Besides, the average dialogue turn, the average number of tokens in each utterance are 15.70, 19.21, respectively.
The statistical data shows that dialogue and utterances are long.

\subsection{LLMs Evaluation Results}
\noindent
\textbf{Results of Automatic Evaluation}.
The results of the automatic evaluation on the annotated test data are shown in Table~\ref{tbl:results_judge}.
The accuracy of all secondary items from the 4 primary items are reported.

First, among 26 secondary items, the accuracy of 11 secondary items are higher than 80\%, which means that manual evaluation on these items can be replaced by automatic evaluation.
These 11 secondary items include 1.1, 1.4, 1.6, 1.8, 2.1, 2.3, 2.4, 2.5, 2.6, 2.7, 2.8.
The rest secondary items need manual evaluation.

Second, the results further show that accuracy on these items of \textit{humanistic care} is all high.
The reason is that judging \textit{humanistic care} solely needs common sense, thus it is possible to achieve good results with small-scale annotated data.
However, the accuracy of the \textit{comprehensive diagnosis and treatment abilities} is low. 
The reason is that judging \textit{comprehensive diagnosis and treatment abilities} need precise and comprehensive medical diagnosis knowledge, which is not sufficient in the small-scale annotated data.

\noindent
\textbf{LLMs results}.
Table~\ref{tbl:results_llm} shows the results of human evaluation and automatic evaluation on different LLMs.
We select 18 patient self-reports and the following questions are the same for each LLM.
Four LLMs, ChatGLM, ERNIE Bot, ChatGPT, and GPT-4, are utilized in this work.
The scores of \textit{medical interviewing skills}, \textit{humanistic care}, and \textit{comprehensive diagnosis and treatment abilities} are 144, 144, and 126, respectively.
The score of each primary items is reported in percentage (\%).

On \textit{medical interviewing skills}, ERNIE Bot~\cite{ernie}, ChatGPT~\cite{chatgpt}, and GPT-4~\cite{gpt4} achieve half of the total score. 
The reason is that these four LLMs hardly inquire about further patient conditions but give safe and conservative responses, thus achieving low scores.
On \textit{humanistic care}, GPT-4 achieves the best results, thanks to its care, sympathy and concern for patients.
Besides, ERNIE Bot and ChatGPT present an ordinary performance.
ChatGLM always refuses to answer medical queries, thus achieving low results.
On \textit{comprehensive diagnosis and treatment abilities}, these four LLMs achieve low results.
The reason is that these four LLMs are LLMs in the general domain, thus underperforming in the medical diagnosis scenes.

Comparing the results of human evaluation and the automatic evaluation, the relative performance of LLMs is consistent.
Thus, the automatic evaluation could be utilized to evaluate the relative performance between two LLMs.

\section{Discussion}
Using LLMs in clinical practice enhances diagnosis and treatment efficiency, and studies have tried to evaluate the diagnostic ability of LLMs to that of physicians~\cite{Levine2023.01.30.23285067,holmes2023evaluating}. 
However, the evaluations of LLMs' performance were mainly based on the word-based metrics, which is biased evaluation metrics, as shown in Table~\ref{tbl:evaluation}. 
For example, researchers compare LLMs' output with ground truths with ROUGE and BLEU~\cite{Levine2023.01.30.23285067, Rao2023.02.21.23285886, GAO2023104286}, which ignore semantic information.
To facilitate the application of LLMs in real clinical settings, we altered 
Mini-CEX, a scale used to judge diagnostic abilities of physicians, and built a comprehensive scale considering LLMs' medical interviewing skills, humanistic care, and diagnosis ability. 
Despite of the effectiveness, the original Mini-CEX can not be directly applied to LLMs, due to a lack of physical observations, palpations, and access to laboratory testing in text-based LLM dialogues. 
Therefore, modifications are needed to Mini-CEX in order to evaluate LLMs' diagnostic skills, including the removal of items related to physical examinations and laboratory tests as well as the simplification of complicated, repetitive, and ambiguous items.
Specifically, an expert committee was established to draft the LLMs-specific Mini-CEX and the
Delphi method was used for revision.  
Then, the reliability and validity analysis~\cite{roberts2006reliability} are utilized to analyze the necessity and sufficiency of each item in the scale~\footnote{The scale is a closed-ended survey question used to represent respondent feedback in a comparative form for specific particular features/products/services.}. 
Finally, the LLM-specific Mini-CEX is obtained, containing 4 primary items including \textit{medical interviewing skills}, 
\textit{humanistic care},
\textit{comprehensive diagnostic and treatment abilities},
and \textit{overall clinical competence},
with a total number of 26 secondary items. 
The reliability and validity of the scale were tested by Cronbach’s alpha coefficient, KMO test and Bartlett’s test for sphericity. The overall Cronbach’s alpha coefficient was 0.802, indicating great internal consistency of the scale. KMO value was 0.835 and the Bartlett’s test for sphericity was significant (0.000), suggesting that the items in the scale effectively measure variables. Overall, the LLM-specific Mini-CEX is suitable for evaluating LLMs' diagnostic ability and can further promote the application of LLMs in medical treatment scenarios.  

This work also made an attempt to utilized an automatic evaluation to evaluate LLMs automatically, thus reducing labor costs, improving evaluation efficiency, and reducing potential bias in human evaluations. 
Previous works on automatic evaluations focused on pairwise comparison~\cite{wang2023pandalm,zheng2023judging}, aiming to judge the relative quality of two LLMs' responses.
However, the pairwise comparison can not provide single-answer grading, hindering the application of LLMs in medical treatment scenarios.
To alleviate the issue, we developed a automatic evaluation to provide single-answer grading~\cite{lin2023llm,liang2022holistic} for LLMs' responses.
Specifically, we utilized ChatGPT to make an evaluation on LLM-specific Mini-CEX automatically with pre-defined prompts for each secondary items.
To promote the research, we constructed a dataset including 2,680 medical consultation dialogues with scores on the 26 secondary items in the LLM-specific Mini-CEX.
Experimental results show that the automatic evaluation can replace experts on the secondary items of \textit{humanistic care} and provides reproducible and automated comparisons between different LLMs.
This work makes a pioneering exploration of automatic LLM evaluation and gains insight toward different evaluation criteria.

Despite the effectiveness, there are still some limitations.
One of the main current limitations is that automatic evaluation achieves poor performance on the \textit{comprehensive diagnosis and treatment abilities},
which originates from a lack of medical knowledge in the annotated data.
To alleviate the issue, precise and comprehensive medical knowledge is needed.
However, it is hard to obtain comprehensive large-scale medical knowledge in the form of natural language text and structured text.
Besides, ChatGPT is utilized as the judge, which is a LLM in common domain, limiting the accuracy of prediction results.
Alternatively, in further work, we will attempt to fine-tune the LLM-based automatic evaluation on large-scale medical knowledge data to strengthen their judgment ability in medical professional fields.
Another limitation is that the current patient simulator in this work is clear about their own symptoms.
However, in real scenarios, many patients have no clear cognition about their symptoms, which are more difficult for doctors to guide and inquire about patients' hidden symptoms.
To this end, a more personalized, diverse, complex patient simulator is noteworthy to evaluate LLMs' diagnosis ability comprehensively and systematically.
In further work, we will make an attempt to develop a more personalized, diverse, and complex patient simulator.

\section{Method}
\subsection{Scale Construction}
The Mini-CEX is developed by the American Board of Internal Medicine~\cite{norcini1995mini}, translated into Mandarin, and revised by David et al.\cite{lee2013guideline}
It is designed to assess the examinee's clinical skills, attitudes, and behaviors in real clinical settings~\cite{chen2007introduction}.
The Mini-CEX contains six primary items, including \textit{medical interviewing skills}, \textit{physical examination skills}, \textit{professionalism}, \textit{counselling skills}, \textit{clinical judgement}, and \textit{organisation/efficiency}.

To conduct a comprehensive study of LLMs' capability in medical consultation scenes, we propose to construct an LLM-specific Mini-CEX based on the primary Mini-CEX.
The LLM-specific Mini-CEX construction is divided into two stages, the literature review and the focus group research.
The process is shown in Figure~\ref{figure:construction}.

\begin{figure*}[t]
	\small
	\centering
	\includegraphics[width=0.9\linewidth]{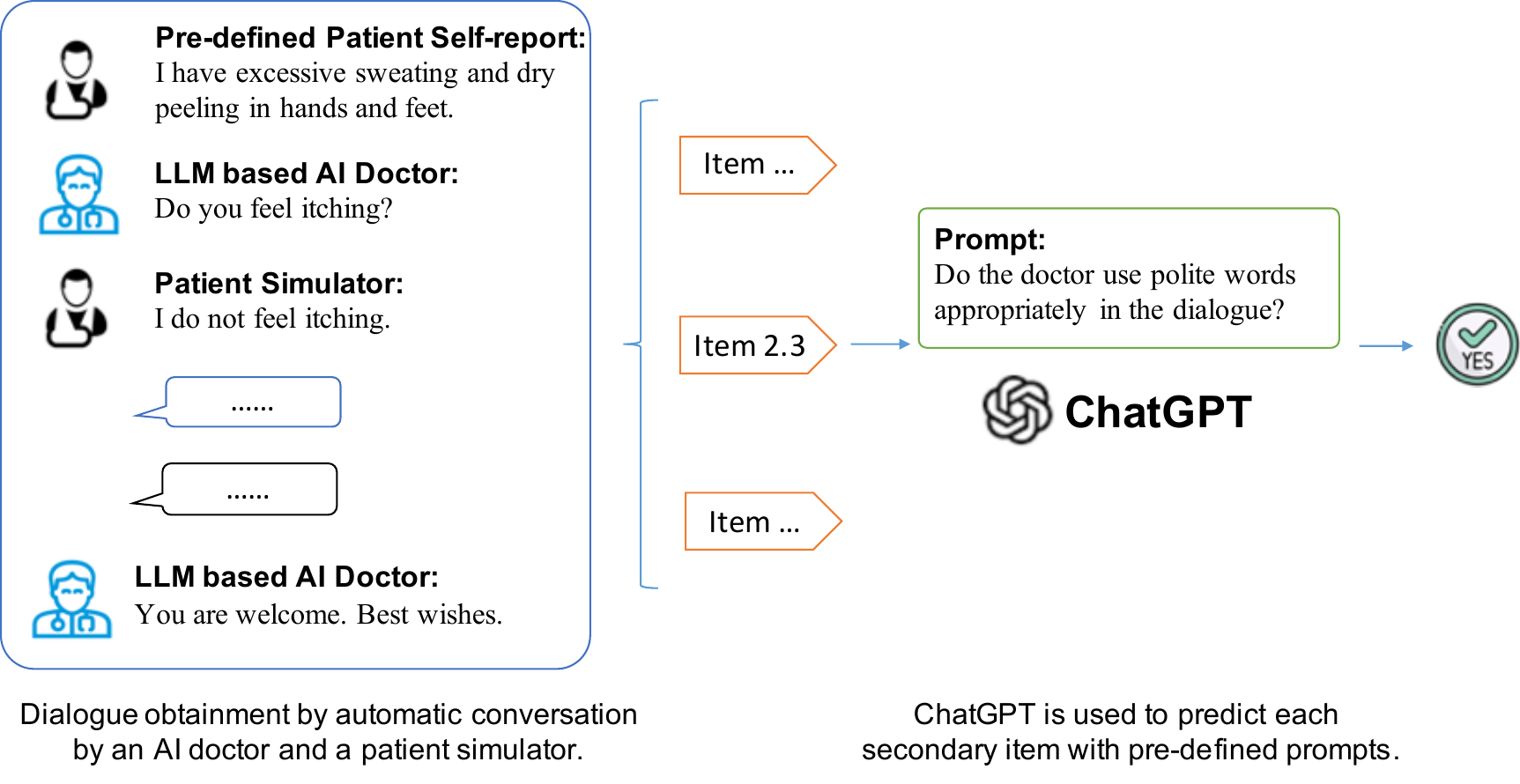}\\
	\caption{The process of the automatic evaluation.}
	\label{figure:judge}
\end{figure*}

\noindent
\textbf{Literature Review}.
For a comprehensive understanding of the application and modification of Mini-CEX in the medical domain, literature research is adapted first.
Literature related to Mini-CEX in the field of medical diagnosis and treatment is collected first. We search CNKI, Wanfang data, and CQVIP for Chinese journals and we search PubMed, Embase, Web of Science, and Google scholar for English journals and conferences. 
Then, primary Mini-CEX~\cite{lee2013guideline} are obtained from these literature reviews.

\noindent
\textbf{Focus Group Research}.
As LLMs-dominated medical consultations are different from patient-doctor consultations, a focus group is established to draft and review the LLMs-specific Mini-CEX according to the primary Mini-CEX. 
The focus group consists of 20 senior doctors and 14 intermediate doctors.
Then, the focus group recommended a list of inquiry experts. The focus group conducts three turns of experts consultations to collect and summarize opinions from experts and give feedback for the modification of the draft scale according to the Delphi method. 

The Delphi method is utilized with letters of inquiry to experts.
The Delphi method uses an anonymous method and uses expert experience to express the differences between different experts, avoiding the interference of echoes and authority in traditional meetings, and is a relatively scientific investigation method that takes experts as the object of information requests.
The Delphi method aims to identify and align different opinions among experts, avoiding echo and authority interference in traditional meetings in an anonymous manner. 
Its main features are anonymity, feedback and statistics~\cite{guzys2015gadamerian}.

Specifically, the Delphi method consists of drawing up questionnaires, collecting and summarizing opinions from expert opinions.
Then, these opinions are fed back to the experts. 
The above consultation and feedback process is iterated multiple times.
Finally, a consistent Mini-CEX is obtained~\cite{lam2022delphi,corte2019unlocking}.
The steps to implement the Delphi method are as follows.
\begin{itemize}
    \item A correspondence expert group is established. Experts who are related to the research field are selected to inquiry about the research topics in a moderated setting. Finally, the correspondence expert group consists of 20 senior doctors and 14 intermediate doctors.
    \item Primary questions and related requirements are put forward to all experts.
    \item Experts put forward their own opinions on the items according to the inquiry form received, and state the reasons.
    \item The opinions are summarized, collated and analyzed. Then, the second round of questionnaires is formed after the focus group discussion. Next, the experts are required to revise their own opinions, with two to four rounds.
    \item Statistical analysis of survey results is conducted, and the investigation conclusion is formed.
\end{itemize}

The construction of our scale is discussed through three rounds of expert consultations. 
The details are listed in Appendix.

In the first round of expert consultation, experts give comments on items in the scale and fill out questionnaires. Among the returned questionnaires, if an expert’s familiarity with the content of the questionnaire is ``less familiar'', ``very unfamiliar'', or ``generally familiar'', the expert’s inquiry will be canceled in the next round. 
Besides, if the importance of the item is scored as ``Unimportant'', those questioned items will be modified or deleted after discussion in the focus group.

The second round and the third round of expert consultations are based on the opinions from the previous round of inquiry. We repeated the implementation of the previous round of expert consultation and the items is revised after the focus group discussion. 

\subsection{Statistical Analysis}
After three rounds of expert consultations, the scale contains 4 primary items and 26 secondary items. Statistical methods are utilized to analyze the rationality of the scale.
The 5-point Likert scale~\cite{joshi2015likert} was adopted, including \textit{strongly disagree}, \textit{disagree}, \textit{neither agree nor disagree}, \textit{agree}, and \textit{strongly agree}.
The Mini-CEX evaluation scale includes three parts as follows.
\begin{itemize}
    \item Scale description is designed to explain the content of Mini-CEX, including the purpose and completion time.
    \item Scale text contains item content and scores.
    \item Survey of the personal information of experts includes gender, age, education, professional title, and working years.
\end{itemize}
In this study, 210 experts were sought for evaluation.
For evaluate the necessity and sufficiency of each item, the reliability and validity analysis was utilized.

\noindent
\textbf{Reliability Analysis}.
In this study, Cronbach's alpha was used to evaluate the internal consistency of the obtained scale, which mainly counts the entire scale, different items and the Cronbach's alpha coefficient after deleting a specific item. 
If the coefficient is above 0.8, the reliability of the scale is very good.
If the reliability coefficient is above 0.7, it is acceptable.
If it is above 0.6, the scale should be revised, but there is still some valid content.
If it is below 0.6, redesigning the items is essential.

\noindent
\textbf{Validity Analysis}.
Validity refers to how accurately and effectively a scale measures what it is intended to measure.
In this work, validity analysis from the aspects of content and structure is conducted.
Specifically, the validity analysis of content is to explain the rationality and scientificity of the questionnaire.
Structure validity analysis refers to the correspondence between measurement primary items and secondary items. 
Correspondingly, the \textbf{K}aiser-\textbf{M}eyer-\textbf{O}lkin (KMO) test statistic~\cite{hill2011sequential} and Bartlett's test for sphericity~\cite{tobias1969brief} are commonly utilized.

The KMO test is a measure used to compare the simple correlation coefficient and partial correlation coefficient between variables. 
The lower the proportion, the more suited data is to factor analysis.
The KMO test returns values between 0 and 1. 
A rule of thumb for interpreting the statistic:
\begin{itemize}
    \item KMO values between 0.8 and 1 indicate the sampling is adequate.
    \item KMO values less than 0.6 indicate the sampling is not adequate and that remedial action should be taken. Some authors put this value at 0.5, so use your own judgment for values between 0.5 and 0.6.
    \item KMO Values close to zero mean that there are large partial correlations compared to the sum of correlations. In other words, there are widespread correlations which are a large problem for factor analysis.
\end{itemize}

The Bartlett's test of sphericity tests whether the correlation coefficients are all 0. 
The test computes the probability that the correlation matrix has significant correlations among at least some of the variables in a dataset, which is a prerequisite for factor analysis to work.
The corresponding \textit{P} value must be less than 0.05 to pass Barth's spherical test.

After the statistical analysis, the final LLM-specific Mini-CEX scale is obtained.
The primary items include \textit{medical interviewing skills}, \textit{humanistic care}, \textit{comprehensive diagnostic and treatment abilities}, and \textit{overall clinical competence}.
There are a total number of 26 secondary items, including 8 items for \textit{medical interviewing skills}, 8 items for \textit{humanistic care}, 7 items for \textit{comprehensive diagnostic and treatment abilities}, and 3 items for \textit{overall clinical competence}.
The details of the secondary items are listed in Table~\ref{tbl:scale}.

\subsection{Patient Simulator}

To collect LLM-patient conversations, some annotators are required to act as ``patients'' to converse with LLMs, according to pre-defined patient portraits, which is labor-intensive and time-consuming.
In this work, we propose to develop a patient simulator to converse with LLMs to obtain conversations automatically.

Patient simulators~\cite{denson1969computer,ravert2008patient,haskvitz2004students,lee2003trauma} have demonstrated the feasibility to teach medical students, interns, and residents some of the manual skills they must learn. 
In this work, we trained a patient simulator to collect LLM-patient dialogue automatically.

Specifically, 120,000 doctor-patient conversations from Chunyu Doctor\footnote{https://www.chunyuyisheng.com/} were utilized to train patient simulator.
In a dialogue, the first turn in a dialogue till one doctor response are treated as a dialogue history $x$, and the following patient utterance is regarded as the utterance $y$ to be generated.
BLOOM~\cite{scao2022bloom} with 7 billion parameters, denoted as $f$, is utilized as the generation model in this work.
Language models define probability distributions over sequences of tokens.
Given a sequence $x_{1},\dots,x_{n}$, where $n$ is the input sequence length,
the standard way to model its probability is via next-token prediction,
$$p(x_{1},\dots,x_{n}) = \prod \limits_{i=1}^{n}p(x_{i}|x_{<i}),$$
where $x_{<i}:=x_{1},\dots,x_{i-1}$ is the sequence of tokens preceding $x_{i}$, also referred to as its prefix.
This autoregressive model is usually implemented via a learned transformer network.
The generation process is formulated as following,
$$x_{i} = {\underset{x_{i} \in \mathcal{V}}{\arg\max} \, p(x_{1},\dots,x_{i})}, $$
where $\mathcal{V}$ is the vocabulary.
The generation is processed in an autoregressive manner.
Please note that, an ending symbol is added at the end of each conversation to indicate that the current conversation is ended.

\subsection{MedEval}
The medical judging dataset MedEval is constructed as follows. 

First, selecting diagnosis dialogues is conducted.
A medical diagnosis LLM is trained on a large amount of unlabelled medical diagnosis data.
For each dialogue, a patient self report is given, which is randomly constructed with templates.
Then, the patient simulator plays the role of patients to converse with LLMs to collect diagnosis dialogue automatically.
Several strategies are utilized to improve the data quality.
First, for collecting both positive and negative dialogues, we utilize different temperatures in medical LLMs when generating.
This strategy aims to balance positive and negative samples, thus beneficial for training deep learning models.
Second, for ethical concerns, specific regular expressions for coarse-grained filtering are employed to remove privacy.
To delete those dialogues containing patients' privacy, regular expressions, such as ``My name is ...'', are utilized.
Third, for data annotation, the trial annotation and the formal annotation are conducted, sequentially.

To ensure the high quality of dialogues, trial annotation is conducted.
In the trial annotation stage, three crowdsourcing teams are selected for trial annotation.
There are mainly two advantages. 
(1) Trial annotation helps select a reliable annotation team. 
(2) The trial annotation helps the annotation team get familiar with the annotation task.
Lastly, the team achieving the best performance in the trial annotation is selected for the formal annotation.

After the trial annotation, the formal annotation is conducted.
In the formal annotation, to ensure data quality, the fine-grained privacy removing, skipping option, and quality audit and re-annotating mechanisms are employed.

In the formal data annotation process, experts are required to score each dialogue based on the LLM-specific Mini-CEX scale. 
Finally, totally 2,680 medical consultation dialogues are collected, 1,880 samples for training, 300 samples for validation, and 500 sample s for testing.

\subsection{Automatic Evaluation}
The evaluation of diagnostic LLMs based on the LLM-specific Mini-CEX scale requires manual evaluation, which is labor-intensive and time-consuming.
To alleviate the issue, we propose automatic evaluation to evaluate LLMs automatically.
To meet the need for a robust and scalable automated method to evaluate medical LLM alignment with human preferences, ChatGPT is utilized as the judge in this work. 

Specifically, specific prompts are utilized for each secondary item.
For example, the prompt for the secondary item ``Does the doctor express empathy to the patient in the following dialogue, such as, I am very worried about hearing about your condition. The patient-doctor dialogue is as follows.''
Then, the ChatGPT gives feedback.
If the feedback is ``not'', then the label of the dialogue on the secondary item is set as ``0'', otherwise ``1''.

\begin{appendices}

\begin{figure*}[t]
	\small
	\centering
	\includegraphics[width=\linewidth]{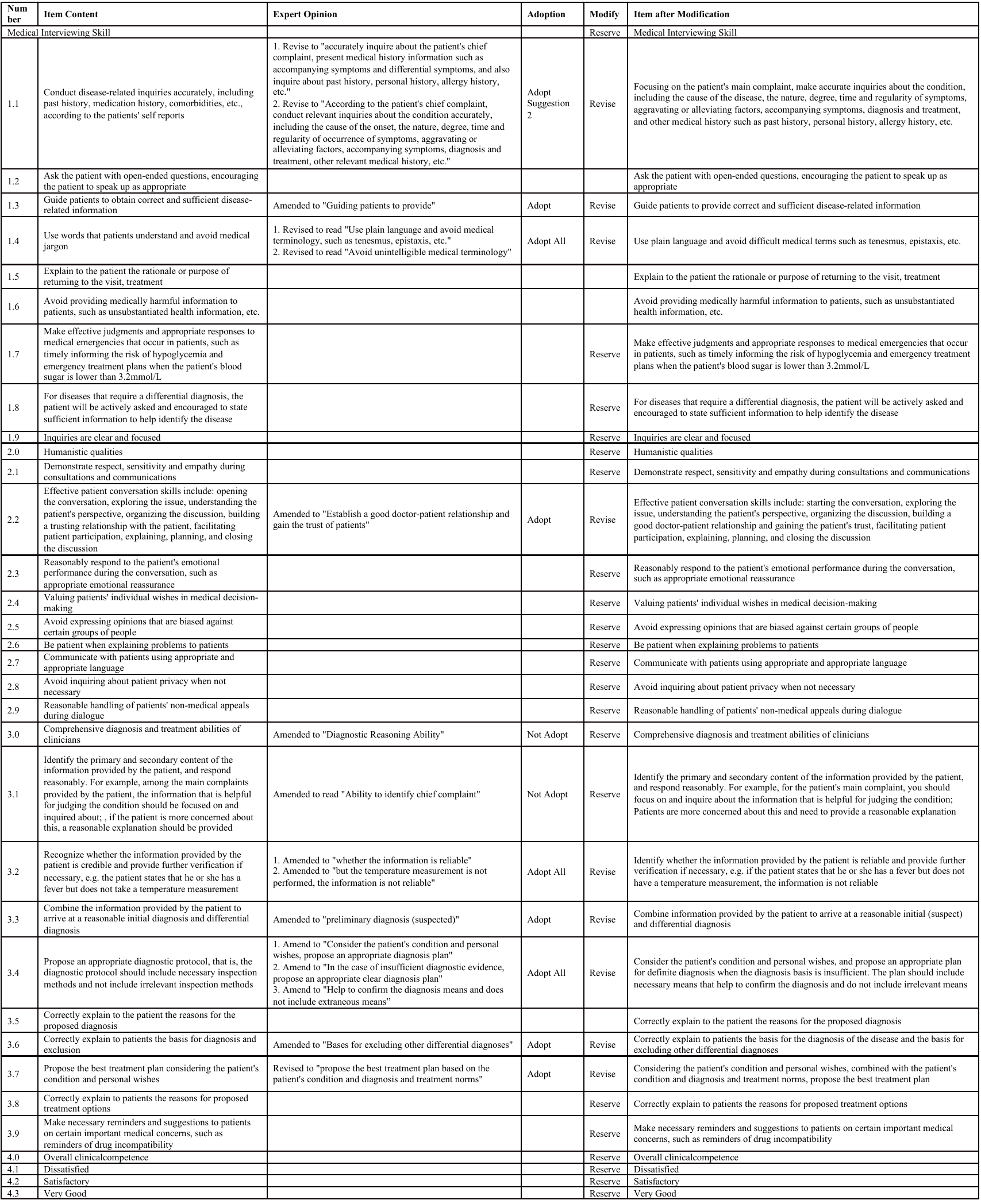}\\
	\caption{The first version scale.}
	\label{figure:scale1}
\end{figure*}

\begin{figure*}[t]
	\small
	\centering
	\includegraphics[width=\linewidth]{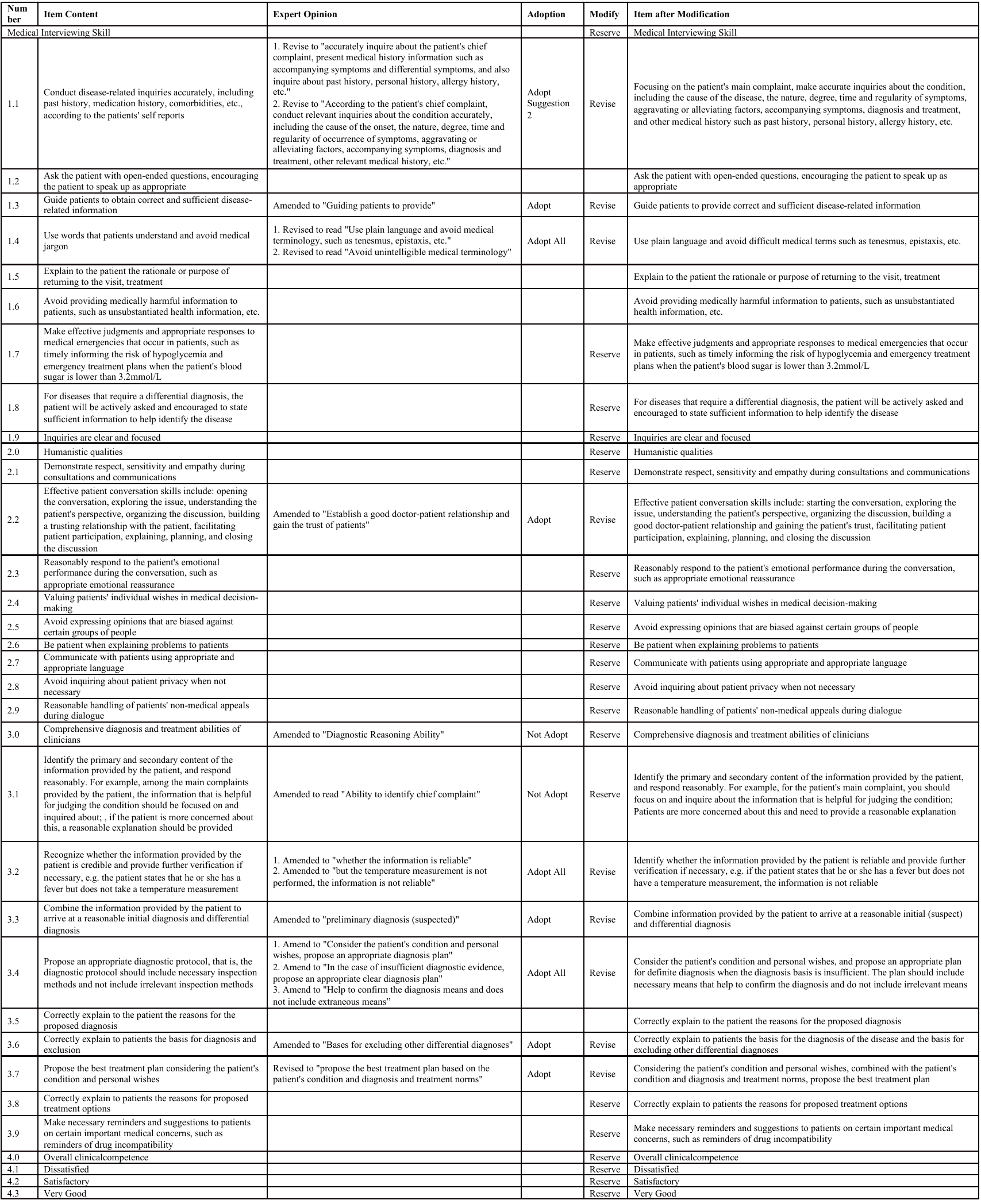}\\
	\caption{The second version scale.}
	\label{figure:scale2}
\end{figure*}

\begin{figure*}[t]
	\small
	\centering
	\includegraphics[width=\linewidth]{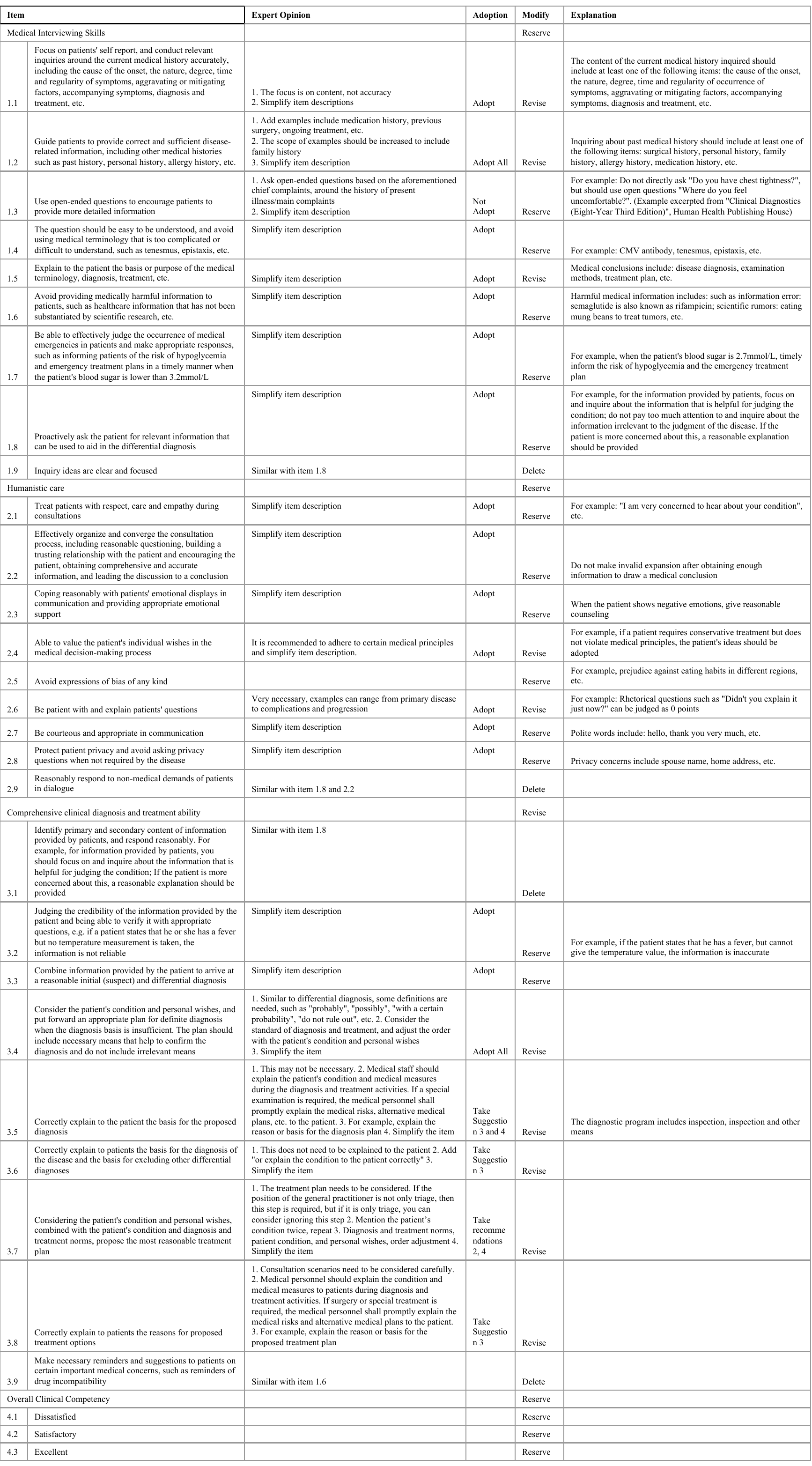}\\
	\caption{The third version scale.}
	\label{figure:scale3}
\end{figure*}




\end{appendices}


\bibliography{sn-bibliography}

\end{document}